\documentclass[11pt]{article}

\usepackage[in]{fullpage}
\usepackage{graphicx,verbatim}
\usepackage[rflt]{floatflt}
\usepackage{epsfig,times,subfigure,graphicx}
\usepackage{amsmath,amssymb,amsopn,algorithm,algorithmic,theorem,float,bbm,bm,enumerate,color,multirow}
\usepackage{fullpage,setspace}
\usepackage{rotating}
\usepackage{array}
\usepackage[bookmarks=false]{hyperref}
\usepackage{url}
\usepackage{chngcntr}
\usepackage{nameref}
\usepackage{zref-xr}
\zxrsetup{toltxlabel}

\newcommand{\beq}{\begin{equation}}
\newcommand{\eeq}{\end{equation}}

\newcommand\s{\mathbb{S}}
\newcommand\R{\mathbb{R}}

\renewcommand\P{\mathbb{P}}

\newcommand{\bX}{\mathbf{X}}

\newcommand{\vertiii}[1]{{\left\vert\kern-0.25ex\left\vert\kern-0.25ex\left\vert #1
    \right\vert\kern-0.25ex\right\vert\kern-0.25ex\right\vert}}

\newcommand{\E}{\mathbb{E}}

\DeclareMathOperator{\argmax}{argmax}
\DeclareMathOperator{\argmin}{argmin}

\DeclareMathOperator{\tr}{Tr}
\DeclareMathOperator{\cone}{cone}
\DeclareMathOperator{\col}{colsp}
\DeclareMathOperator{\row}{rowsp}
\DeclareMathOperator{\rank}{rank}

\DeclareMathOperator{\Diag}{Diag}
\DeclareMathOperator{\tn}{tr}
\DeclareMathOperator{\sk}{sk}
\DeclareMathOperator{\op}{op}

\newcounter{exampleI}
{\theorembodyfont{\rmfamily} \theoremstyle{plain} }

\newcounter{exampleII}
{\theorembodyfont{\rmfamily} \theoremstyle{plain} }

\newcounter{exampleIII}
{\theorembodyfont{\rmfamily} \theoremstyle{plain} }

{\theorembodyfont{\rmfamily} }
{\theorembodyfont{\rmfamily} \newtheorem{defn}{Definition}}
\newtheorem{theo}{Theorem}

\newtheorem{lemm}{Lemma}

\newcommand{\proof}{\noindent{\itshape Proof:}\hspace*{1em}}

\newcommand{\qed}{\nolinebreak[1]~~~\hspace*{\fill} \rule{5pt}{5pt}\vspace*{\parskip}\vspace*{1ex}}

\newcommand {\commentout}[1] {}

\def\ints{{{\rm Z} \kern -.35em {\rm Z} }}  
\def\smallints{{{\rm Z} \kern -.3em {\rm Z} }}  
\def\pints{{{\rm I} \kern -.15em {\rm N} }}      
\newcommand{\reals}{\mathbb R}

\def\cplx{{{\rm I} \kern -.45em {\rm C} }}       
\def\l2{\rm {\mathcal L}^{2}(\reals)}            

\newtheorem{nad}{Notation and Definitions}[section]

\newtheorem{theorem}{Theorem}[section]

\newcommand{\be}{\begin{eqnarray}}
\newcommand{\ee}{\end{eqnarray}}
\newcommand{\bea}{\begin{eqnarray}}
\newcommand{\eea}{\end{eqnarray}}
\newcommand{\beaa}{\begin{eqnarray*}}
\newcommand{\eeaa}{\end{eqnarray*}}
\newcommand{\bnad}{\begin{nad}}
\newcommand{\enad}{\end{nad}}

\newcommand{\diam}{{\rm diam\,}}

\title{Structured Matrix Recovery via the Generalized Dantzig Selector}
\date{\today}
\author{Sheng Chen \qquad \qquad Arindam Banerjee \vspace*{2mm}
\\
\{shengc,banerjee@cs.umn.edu\}\vspace*{2mm}\\
Department of Computer Science \& Engineering\\
University of Minnesota, Twin Cities}

\begin{document}

\maketitle

\begin{abstract}
In recent years, structured matrix recovery problems have gained considerable attention for its real world applications, such as recommender systems and computer vision. Much of the existing work has focused on matrices with low-rank structure, and limited progress has been made matrices with other types of structure. In this paper we present non-asymptotic analysis for estimation of generally structured matrices via the generalized Dantzig selector under generic sub-Gaussian measurements. We show that the estimation error can always be succinctly expressed in terms of a few geometric measures of suitable sets which only depend on the structure of the underlying true matrix. In addition, we derive the general bounds on these geometric measures for structures characterized by unitarily invariant norms, which is a large family covering most matrix norms of practical interest. Examples are provided to illustrate the utility of our theoretical development. 
\end{abstract}

\section{Introduction}
\label{sec:intro}
Structured matrix recovery has found a wide spectrum of applications in real world, e.g., recommender systems \cite{kobv09}, face recognition \cite{clmw11}, etc. The recovery of an unknown structured matrix $\Theta^* \in \R^{d \times p}$ essentially needs to consider two aspects: the measurement model, i.e., what kind of information about the unknown matrix is revealed from each measurement, and the structure of the underlying matrix, e.g., sparse, low-rank, etc. In the context of structured matrix estimation and recovery, a widely used measurement model is the linear measurement, i.e., one has access to $n$ observations of the form
\beq
y_i = \langle \langle \Theta^*, X_i \rangle \rangle + \omega_i~,
\eeq
for $\Theta^*$, where $\langle \langle \cdot, \cdot \rangle \rangle$ denotes the matrix inner product, i.e., $\langle \langle A, B \rangle \rangle = \tr(A^T B)$ for any $A, B \in \R^{d \times p}$, and $\omega_i$'s are additive noise. In the literature, various types of measurement matrices $X_i$ has been investigated, for example, Gaussian ensemble where $X_i$ consists of i.i.d. standard Gaussian entries \cite{crpw12}, rank-one projection model where $X_i$ is randomly generated with constraint $\rank(X_i) = 1$ \cite{cazh15}. A special case of rank-one projection is the matrix completion model \cite{care09}, in which $X_i$ has a single entry equal to 1 with all the rest set to 0, i.e., $y_i$ takes the value of one entry from $\Theta^*$ at each measurement. Other measurement models include row-and-column affine measurement \cite{zuwa15}, exponential family matrix completion \cite{gurg14,gubg15}, etc. 

Previous work has shown 
that low-complexity structure of $\Theta^*$, often captured by a small value of some norm $R(\cdot)$, can significantly benefit its recovery \cite{crpw12,nrwy12}. For instance, one of the popular structures of $\Theta^*$ is low-rank, which can be approximated by a small value of the trace norm $\| \cdot \|_{\tn}$. Under the low-rank assumption of $\Theta^*$, numerous recovery guarantees  
have been established for different measurement matrices using convex programs, e.g., trace-norm regularized least-square estimator \cite{capl09,refp10,nrwy12,gurg14},
\begin{align}
\label{eq:mat_lasso}
\underset{\Theta \in \R^{d \times p}}{\min} \ \ \frac{1}{2} \sum_{i=1}^{n} \left(y_i - \langle \langle X_i, \Theta \rangle \rangle\right)^2 + \beta_n \|\Theta^*\|_{\tn} ~,
\end{align}
and constraint trace-norm minimization estimators \cite{capl09,refp10,crpw12,cazh15,gubg15}, 
\begin{gather}
\label{eq:bp}
\underset{\Theta \in \R^{d \times p}}{\min} \ \|\Theta\|_{\tn} \quad \text{s.t.} \quad \sum_{i=1}^{n} \left(y_i - \langle \langle X_i, \Theta \rangle \rangle \right)^2 \leq \epsilon_n^2  ~, \\
\label{eq:l1_dantzig}
\underset{\Theta \in \R^{d \times p}}{\min} \ \|\Theta\|_{\tn} \quad \text{s.t.} \quad \| \sum_{i=1}^n \left(\langle \langle X_i, \Theta \rangle \rangle - y_i \right) X_i \|_{\op} \leq \lambda_n ~,
\end{gather}
where $\beta_n$, $\epsilon_n$, $\lambda_n$ are tuning parameters, and $\|\cdot\|_{\op}$ denotes the operator (spectral) norm.
Among the convex approaches, the exact recovery guarantee of constraint estimator \eqref{eq:bp} was analyzed for the noiseless setting in \cite{refp10}, under certain matrix-form restricted isometry property (RIP). In the presence of noise, \cite{capl09} also used matrix RIP to establish the recovery error bound for both regularized and constraint estimators, i.e., both \eqref{eq:mat_lasso} and \eqref{eq:l1_dantzig}. In \cite{cazh15}, a variant of estimator \eqref{eq:l1_dantzig} was proposed and its recovery guarantee was built on a so-called restricted uniform boundedness (RUB) condition, which is more suitable for the rank-one projection based measurement model. Despite the fact that the low-rank structure has been well studied, only a few works extend to more general structures. In \cite{nrwy12}, the regularized estimator \eqref{eq:mat_lasso} was generalized by replacing the trace norm with a decomposable norm $R(\cdot)$ for other structures. \cite{crpw12} aimed at constraint estimator \eqref{eq:bp} with $\|\cdot\|_{\tn}$ replaced by a norm from a broader class called atomic norm, but the consistency of the estimator is only available when the noise vector is bounded. In matrix completion setting, recovery guarantee were also analyzed for general norms in \cite{gurg14,gubg15}. 

In this work, we present a general framework for estimation of structured matrices via the generalized Dantzig sector (GDS) \cite{chcb14,calr14} as follows
\beq
\label{eq:dantzig}
\begin{gathered}
\hat{\Theta} = \underset{\Theta \in \R^{d \times p}}{\argmin} \ \ R(\Theta) \ \ \ \text{s.t.} \ \ \ R^* \left( \sum_{i=1}^n \left(\langle \langle X_i, \Theta \rangle \rangle - y_i \right) X_i \right) \leq \lambda_n ~,
\end{gathered}
\eeq
in which $R(\cdot)$ can be arbitrary norm and its dual norm is $R^*(\cdot)$. Note that the estimator \eqref{eq:l1_dantzig} is a special case of the formulation above, as operator norm is dual to trace norm. Our deterministic analysis of the estimation error $\| \hat{\Theta} - \Theta^* \|_F$ relies on a condition based on a suitable choice of $\lambda_n$ and the restricted strong convexity (RSC) condition \cite{nrwy12,bcfs14}. By assuming sub-Gaussian $X_i$ and $\omega_i$, we show that these conditions are satisfied with high probability, and the recovery error can be expressed in terms of certain \emph{geometric measures} of sets associated with $\Theta^*$. Such a geometric characterization is inspired by related advances in recent years \cite{nrwy12,crpw12,bcfs14}. One key ingredient in such characterization is the \emph{Gaussian width} \cite{gord85}, which measures the size of sets in $\R^{d \times p}$. Related advances can be found in \cite{crpw12,chcb14,calr14}, but they all rely on the measurements being a Gaussian ensemble, to which classical concentration results \cite{gord85,gord88} are directly applicable. In contrast, our work allows general sub-Gaussian measurement matrices and noise, by suitably using ideas from generic chaining \cite{tala05,tala14}, a powerful geometric approach to bounding stochastic processes.
Our results can also be extended to heavy tailed designs and noise, following recent advances \cite{sibr15}. 
From a practical viewpoint, we derive the general bounds of those geometric measures for the class of \emph{unitarily invariant} norms. By its name, this class of matrix norms is invariant under any unitary transformation, i.e., for any matrix $\Theta \in \R^{d \times p}$, its norm value is equal to that of $U \Theta V$ if both $U \in \R^{d \times d}$ and $V \in \R^{p \times p}$ are unitary matrices. The widely-used trace norm, spectral norm and Frobenius norm all belong to this class. A well-known result established in \cite{vonn37} is that any unitarily invariant matrix norm is equivalent to some vector norm applied on the set of singular values (see Lemma \ref{lem:uni_inv} for details), and this equivalence allows us to use the techniques developed in \cite{chba15} for vector norms to derive the bounds of the geometric measures for unitarily invariant norms.
We illustrate concrete versions of the general bounds using the trace norm and the recently proposed spectral $k$-support norm \cite{mcps14}.

The rest of the paper is organized as follows: we first provide the deterministic analysis in Section \ref{sec:det_analysis}. In Section \ref{sec:prelim}, we introduce the preliminaries of some probability tools, which are used in the later analysis. In Section \ref{sec:rand_analysis}, we present the probabilistic analysis for sub-Gaussian measurement matrices and noise, along with the general bounds of the geometric measures for unitarily invariant norms. Section \ref{sec:example} is dedicated to the examples for the application of general bounds, and we conclude in Section \ref{sec:conc}.

\section{Deterministic Recovery Guarantees}
\label{sec:det_analysis}
To evaluate the performance of the estimator \eqref{eq:dantzig}, we mainly focus on the Frobenius-norm error, i.e., $\|\hat{\Theta} - \Theta^* \|_{F}$. Throughout the paper, w.l.o.g. we assume that $d \leq p$. For convenience, we denote the collection of $X_i$'s by $\bX = \{ X_i \}_{i=1}^n$, and let $\omega = [\omega_1, \omega_2, \ldots, \omega_n]^T$ represent the noise vector. In the following theorem, we provide a deterministic bound for $\|\hat{\Theta} - \Theta^*\|_{F}$ under some standard assumptions on $\lambda_n$ and $\bX$.

\begin{theo}
\label{theo:dantzig}
Define the set
\begin{align*}
\mathcal{E}_R(\Theta^*) = \cone \{ \Delta \ | \ R(\Delta + \Theta^*) \leq R(\Theta^*) \} ~.
\end{align*}
Assume the following conditions hold for $\lambda_n$ and $\bX$,
\beq
\label{eq:lambda_cond}
\lambda_n \geq R^*\left( \sum_{i=1}^n \omega_i X_i \right) ~,
\eeq
\beq
\label{eq:re_cond}
\frac{\sum_{i=1}^{n} \langle \langle X_i, \Delta \rangle \rangle^2}{\|\Delta\|_{F}^2} \geq \alpha > 0, \ \forall \ \Delta \in \mathcal{E}_R(\Theta^*) ~.
\eeq
Then the estimation $\|\hat{\Theta} - \Theta^* \|_{F}$ error satisfies
\beq
\label{eq:gds_bound}
\|\hat{\Theta} - \Theta^*\|_{F} \leq \frac{2 \Psi_{R}(\Theta^*) \cdot \lambda_n}{\alpha} ~,
\eeq
where $\Psi_{R}(\cdot)$ is the restricted compatibility constant defined as
\beq
\label{eq:rcc}
\Psi_{R}(\Theta^*) = \sup_{\Delta \in \mathcal{E}_R(\Theta^*)} \frac{R(\Delta)}{\|\Delta\|_{F}}~.
\eeq
\end{theo}
\proof Since $\lambda_n$ satisfies the condition \eqref{eq:lambda_cond} and $\omega_i = y_i - \langle \langle X_i, \Theta^* \rangle \rangle$, we have
\begin{align*}
R^*\left( \sum_{i=1}^n \left(\langle \langle X_i, \Theta^* \rangle \rangle - y_i \right) X_i  \right) \leq \lambda_n ~,
\end{align*}
which indicates that the constraint set in \eqref{eq:dantzig} is feasible, thus
\begin{align*}
R^*\left( \sum_{i=1}^n \left(\langle \langle X_i, \hat{\Theta} \rangle \rangle - y_i \right) X_i  \right) \leq \lambda_n ~.
\end{align*}
Using triangular inequality, one has
\begin{align*}
R^* \left( \sum_{i=1}^n \langle \langle X_i, \hat{\Theta} - \Theta^* \rangle \rangle \cdot X_i \right) \leq 2 \lambda_n ~.
\end{align*}
Denote $\hat{\Theta} - \Theta^*$ by $\Delta$, and by the definition of dual norm, we get
\begin{align*}
\sum_{i=1}^n \langle \langle X_i, \Delta \rangle \rangle^2 = \langle \langle \Delta, \sum_{i=1}^n \langle \langle X_i, \Delta \rangle \rangle \cdot X_i \rangle \rangle \leq R(\Delta) \cdot R^* \left( \sum_{i=1}^n \langle \langle X_i, \hat{\Theta} - \Theta^* \rangle \rangle \cdot X_i \right) \leq 2 \lambda_n R(\Delta) ~.
\end{align*}
On the other hand, the objective function in \eqref{eq:dantzig} implies that $R(\hat{\Theta}) \leq R(\Theta^*)$. Therefore the error vector $\Delta$ must belong to the set $\mathcal{E}_R(\Theta^*)$. Using condition \eqref{eq:re_cond}, we obtain
\begin{gather*}
\alpha \| \Delta \|_{F}^2 \leq \sum_{i=1}^n \langle \langle X_i, \Delta \rangle \rangle^2 \leq 2 \lambda_n R(\Delta) ~, \\
\|\Delta\|_{F} \leq \frac{2 \lambda_n}{\alpha} \frac{R(\Delta)}{\|\Delta\|_{F}} \leq \frac{2 \Psi_{R}(\Theta^*) \cdot \lambda_n}{\alpha} ~,
\end{gather*}
which complete the proof. \qed

The convex cone $\mathcal{E}_R(\Theta^*)$ plays a important role in characterizing the error bound, and its geometry is determined by $R(\cdot)$ and $\Theta^*$. The recovery bound assumes no knowledge of the norm $R(\cdot)$ and true matrix $\Theta^*$, thus allowing general structures. In this work, we are particularly interested in $R(\cdot)$ from the class of \emph{unitarily invariant} matrix norm, which essentially satisfies the following property,
\begin{gather}
R(\Theta) = R(U\Theta V)
\end{gather}
for any $\Theta \in \R^{d \times p}$ and unitary matrices $U \in \R^{d \times d}$, $V \in \R^{p \times p}$ ~.
A useful result for unitarily invariant norm is given in the lemma below (see \cite{vonn37,lewi95,bhat97} for details).
\begin{lemm}
\label{lem:uni_inv}
Suppose that the singular values of a matrix $\Theta \in \R^{d \times p}$ are given by $\sigma = [\sigma_1, \sigma_2, \ldots, \sigma_d]^T$. A unitarily invariant norm $R : \R^{d \times p} \mapsto \R$ can be characterized by some symmetric gauge function\footnote{Symmetric gauge function is a norm on $\R^d$ that is invariant under sign-changes and permutations of the elements.} $f : \R^d \mapsto \R$ as
\beq
R(\Theta) = f(\sigma) ~,
\eeq
and its dual norm is given by
\beq
R^*(\Theta) = f^*(\sigma) ~.
\eeq
\end{lemm}
As the sparsity of $\sigma$ equals the rank of $\Theta$, the class of unitarily invariant matrix norms is useful in structured low-rank matrix recovery and includes many widely used norms, e.g., trace norm with $f(\cdot) = \|\cdot\|_1$, Frobenius norm with $f(\cdot) = \|\cdot\|_2$, Schatten $p$-norm with $f(\cdot) = \|\cdot\|_p$, Ky Fan $k$-norm when $f(\cdot)$ is the $\ell_1$ norm of the largest $k$ elements in magnitude, etc.

Before proceeding with the analysis, we introduce some notations. For the rest of paper, we denote by $\sigma(\Theta) \in \R^d$ the vector of singular values (sorted in descending order) of matrix $\Theta \in \R^{d \times p}$, and may use the shorthand $\sigma^*$ for $\sigma(\Theta^*)$. For any $\theta \in \R^d$, we define the corresponding $|\theta|^{\downarrow}$ by arranging the absolute values of elements of $\theta$ in descending order. Given any matrix $\Theta \in \R^{d \times p}$ and subspace $\mathcal{M} \subseteq \R^{d \times p}$, we denote by $\Theta_{\mathcal{M}}$ the orthogonal projection of $\Theta$ onto $\mathcal{M}$. Besides we let $\col(\Theta)$ ($\row(\Theta)$) be the subspace spanned by columns (rows) of $\Theta$. The notation $\s^{dp-1}$ represents the unit sphere of $\R^{d \times p}$, i.e., the set $\{ \Theta |  \|\Theta\|_F = 1 \}$. The unit ball of norm $R(\cdot)$ is denoted by $\Omega_R = \{ \Theta | R(\Theta) \leq 1 \}$. Throughout the paper, the symbols $c, C, c_0, C_0$, etc., are reserved for universal constants, which may be different at each occurrence.

In the rest of our analysis, we will frequently use the so-called ordered weighted $\ell_1$ (OWL) norm for $\R^d$ \cite{fino16}, which is defined as
\beq
\|\theta\|_{w} \triangleq \langle |\theta|^{\downarrow}, |w|^{\downarrow} \rangle ~,
\eeq
where $w \in \R^d$ is a predefined weight vector. Noting that the OWL norm is a symmetric gauge, we define the \emph{spectral OWL norm} for $\Theta$ as: $\|\Theta\|_w \triangleq \|\sigma(\Theta)\|_w$, i.e., by applying the OWL norm on $\sigma(\Theta)$.

\section{Background and Preliminaries}
\label{sec:prelim}
The tools for our probabilistic analysis include the notion of Gaussian width \cite{gord85,gord88}, certain properties of sub-Gaussian random matrices, and generic chaining \cite{tala05,tala14}. Here we briefly introduce the basic ideas and results for each of them as needed for our analysis.

\subsection{Gaussian width}
The Gaussian width can be defined for any subset $\mathcal{A} \subseteq \R^{d \times p}$ as follows \cite{gord85,gord88}, 
\beq
w(\mathcal{A}) \triangleq \E_G \left[ \sup_{Z \in \mathcal{A}} ~\langle \langle G, Z \rangle \rangle \right] ~,
\eeq
where $G$ is a random matrix with i.i.d.~standard Gaussian entries, i.e., $G_{ij} \sim N(0, 1)$. In particular, sometimes we want to upper bound the Gaussian width for a subset of unit sphere, i.e., $\mathcal{A} \subseteq \s^{dp-1}$. A useful inequality \cite{crpw12,almt14} is given by 
\beq
\label{eq:stat_dim}
w^2(\mathcal{A}) \leq \E_G \left[ \inf_{Z \in \mathcal{N}} \| G - Z\|_F^2 \right] ~,
\eeq
in which $\mathcal{N}$ is the polar cone of $\cone(\mathcal{A})$. The quantity on the right-hand side is also called \emph{statistical dimension} of $\cone(\mathcal{A})$ \cite{almt14}.

\subsection{Sub-Gaussian random matrices}
Analogous to sub-Gaussian random vector, a random matrix $X$ is sub-Gaussian with $\vertiii{X}_{\psi_2} \leq \kappa$ if
\beq
\vertiii{\langle \langle X, Z \rangle \rangle}_{\psi_2} \leq \kappa \text{\ \ for any $Z \in \s^{dp-1}$} ~,
\eeq
where the $\psi_2$ norm for sub-Gaussian random variable $x$ is defined as $\vertiii{x}_{\psi_2} = \sup_{q \geq 1} q^{-\frac{1}{2}} (\E |x|^q)^{\frac{1}{q}}$ (see \cite{vers12} for more details of $\psi_2$ norm). One nice property of sub-Gaussian random variable is the thin tail, i.e.,
\beq
\P ( |x| > \epsilon) \leq e \cdot \exp \left( - c \epsilon^2 / \|x\|^2_{\psi_2} \right) ~.
\eeq
To facilitate the computation of Gaussian width, we might use some properties specific to the Gaussian random matrix $G \in \R^{d \times p}$, which are summarized as follows. The symbol ``$\sim$'' means ``has the same distribution as''. 


{\bf Property 1:} Given an $m$-dimensional subspace $\mathcal{M} \subseteq \R^{d \times p}$ spanned by orthonormal basis $U_1, \ldots, U_m$,
\begin{align*}
G_{\mathcal{M}} \sim \sum_{i=1}^{m} g_i U_i,
\end{align*}
where $g_i$'s are i.i.d. standard Gaussian random variables. Moreover, $\E \left[\|G_{\mathcal{M}}\|_F^2\right] = m$. 

\proof Given the orthonormal basis $U_1, \ldots, U_m$ of subspace ${\mathcal{M}}$, $G_{\mathcal{M}}$ can be written as
\begin{align*}
G_{\mathcal{M}} = \sum_{i=1}^m \langle \langle G, U_i \rangle \rangle \cdot U_i
\end{align*} 
Since $\|U_1\|_F = \ldots = \|U_m\|_F = 1$, each $\langle \langle G, U_i \rangle \rangle$ is standard Gaussian. Moreover, as $U_1, \ldots, U_m$ are orthogonal, $\langle \langle G, U_i \rangle \rangle$ are independent of each other. \qed

{\bf Property 2:} $G_{\mathcal{M}_1}$ and $G_{\mathcal{M}_2}$ are independent if $\mathcal{M}_1, \mathcal{M}_2 \subseteq \R^{d \times p}$ are  orthogonal subspaces. 

\proof Suppose that the orthonormal bases of $\mathcal{M}_1, \mathcal{M}_2$ are given by  $U_1, \ldots, U_{m_1}$ and $V_1, \ldots, V_{m_2}$ respectively. Using Property 1 above, $G_{\mathcal{M}_1}$ and $G_{\mathcal{M}_2}$ can be written as
\begin{align*}
G_{\mathcal{M}_1} = \sum_{i=1}^{m_1} \langle \langle G, U_i \rangle \rangle \cdot U_i \ \sim \  \sum_{i=1}^{m_1} g_i U_i ~, \\
G_{\mathcal{M}_2} = \sum_{i=1}^{m_2} \langle \langle G, V_i \rangle \rangle \cdot V_i \ \sim \ \sum_{i=1}^{m_2} h_i V_i ~,
\end{align*}
where $g_1, \ldots, g_{m_1}$ and $h_1, \ldots, h_{m_2}$ are all standard Gaussian. As $\mathcal{M}_1, \mathcal{M}_2 \subseteq \R^{d \times p}$ are orthogonal,  $U_1, \ldots, U_{m_1}$ and $V_1, \ldots, V_{m_2}$ are orthogonal to each other as well, which implies that $g_1, \ldots, g_{m_1}$ and $h_1, \ldots, h_{m_2}$ are all independent. Therefore $G_{\mathcal{M}_1}$ and $G_{\mathcal{M}_2}$ are independent. \qed

{\bf Property 3:} Given a subspace
\begin{align*}
\mathcal{M} = \{ \Theta \in \R^{d \times p} \ | \ \col(\Theta) \subseteq \mathcal{U}, \ \row(\Theta) \subseteq \mathcal{V}\} ~,
\end{align*}
where $\mathcal{U} \subseteq \R^d$, $\mathcal{V} \subseteq \R^p$ are two subspaces of dimension $m_1$ and $m_2$ respectively, then  $\|G_{\mathcal{M}}\|_{\op}$ satisfies
\begin{align*}
\|G_{\mathcal{M}}\|_{\op} \sim \|G'\|_{\op} ~,
\end{align*}
where $G'$ is an $m_1 \times m_2$ matrix with i.i.d. standard Gaussian entries. 

\proof Suppose that the orthonormal bases for $\mathcal{U}$ and $\mathcal{V}$ are $U = [u_1, \ldots, u_{m_1}]$ and $V = [v_1, \ldots, v_{m_2}]$ respectively, and $U_{\perp}$ and $V_{\perp}$ denote the orthonormal bases for their orthogonal complement. It is easy to see that the orthonormal basis for $\mathcal{M}$ can be given by $\{ u_i v_j^T  \  |  \ 1 \leq i \leq m_1, \ 1 \leq j \leq m_2 \}$. Using Property 1, we have
\begin{align*}
G_{\mathcal{M}} \ \sim \ \sum_{i=1}^{m_1} \sum_{j=1}^{m_2} g'_{ij} u_i v_j^T = U G' V = [U, U_{\perp}] \cdot \left[ \begin{array}{ll} G' & 0_{m_1 \times (p - m_2)} \\  0_{(d - m_1) \times m_2} & 0_{(d-m_1) \times (p - m_2)} \end{array} \right] \cdot \left[ \begin{array}{ccc} V^T \\ V_{\perp}^T \end{array} \right]
\end{align*}
where $G'$ is a $m_1 \times m_2$ standard Gaussian random matrix. Note that both $[U, U_{\perp}] \in \R^{d \times d}$ and $[V, V_{\perp}] \in \R^{p \times p}$ are unitary matrices, because they form the orthonormal bases for $\R^d$ and $\R^p$ respectively. If we denote $\left[ \begin{array}{cc} G' & 0 \\  0 & 0 \end{array} \right]$ by $W$, then $\|G_{\mathcal{M}}\|_{\op} = \|W\|_{\op}$ as spectral norm is unitarily invariant. Further, if the SVD of $G'$ is $G' = U_1 \Sigma_1 V_1^T$, where $U_1 \in \R^{m_1 \times m_1}$, $\Sigma_1 \in \R^{m_1 \times m_2}$ and $V_1 \in \R^{m_2 \times m_2}$, then the SVD of $W$ is given by
\begin{align*}
W =  \left[ \begin{array}{ll} U_1 & 0_{m_1 \times (d - m_1)} \\ 0_{(d - m_1) \times m_1} & U_2 \end{array} \right] \left[ \begin{array}{ll} \Sigma_1 & 0_{m_1 \times (p - m_2)} \\  0_{(d - m_1) \times m_2} & 0_{(d-m_1) \times (p - m_2)} \end{array} \right]  \left[ \begin{array}{ll} V_1^T & 0_{m_2 \times (p - m_2)} \\ 0_{(p - m_2) \times m_2} & V_2^T \end{array} \right] ~,
\end{align*}
where $U_2 \in \R^{(d-m_1) \times (d-m_1)}$ and $V_2 \in \R^{(p-m_2) \times (p-m_2)}$ are arbitrary unitary matrices. From the equation above, we can see that $W$ and $G'$ share the same singular values, thus $\|G_{\mathcal{M}}\|_{\op} = \|W\|_{\op} = \|G'\|_{\op}$. \qed
 
{\bf Property 4:} The operator norm $\|G\|_{\op}$ satisfies
\begin{gather}
\label{prop4_1}
\P \left( \|G\|_{\op} \geq \sqrt{d} + \sqrt{p} + \epsilon \right) \leq \exp\left(-\frac{\epsilon^2}{2}\right) ~, \\
\label{prop4_2}
\E\left[\|G\|_{\op}\right] \leq \sqrt{d} + \sqrt{p} ~, \\
\label{prop4_3}
\E\left[\|G\|^2_{\op}\right] \leq \left(\sqrt{d} + \sqrt{p} \right)^2 + 2 ~.
\end{gather}
\proof \eqref{prop4_1} and \eqref{prop4_2} are the classical results on the extreme singular value of Gaussian random matrix \cite{ruve10,vers12} (see Theorem 5.32 and Corollary 5.35 in \cite{vers12}). \eqref{prop4_3} is used in \cite{crpw12} (see (82) - (87) in \cite{crpw12}). \qed

\subsection{Generic Chaining}
Generic chaining is a powerful tool for bounding the supreme of stochastic processes \cite{tala05,tala14}.
Suppose $\{Z_t\}_{t \in \mathcal{T}}$ is a centered stochastic process, where each $Z_t$ is a centered random variable. We assume the index set $\mathcal{T}$ is endowed with some metric $s(\cdot, \cdot)$. In order to use generic chaining bound, the critical condition that $\{Z_t\}_{t \in \mathcal{T}}$ has to satisfy is that, for any $u, v \in \mathcal{T}$,
\beq
\P \left(| Z_u - Z_v| \geq \epsilon \right) \leq c_1 \cdot \exp \left( - \frac{c_2 \epsilon^2}{s^2(u, v)} \right) ~,
\eeq
where $c_1$ and $c_2$ are universal constants. Under this condition, the following results hold for $\{Z_t\}_{t \in \mathcal{T}}$,
\beq
\label{eq:gc_bound}
\E \left[ \sup_{t \in \mathcal{T}} Z_t \right] \leq c_0 \gamma_2 \left(\mathcal{T}, s\right) ~,
\eeq
\beq
\label{eq:gc_conc}
\begin{gathered}
\P \left(\sup_{u,v \in \mathcal{T}} |Z_u - Z_v | \geq C_1 \left(\gamma_2(\mathcal{T}, s) + \epsilon \cdot \diam(\mathcal{T},s) \right) \right) \leq C_2 \exp \left( -\epsilon^2 \right) ~,
\end{gathered}
\eeq
where $\diam(\mathcal{T},s)$ is the diameter of set $\mathcal{T}$ w.r.t. the metric $s(\cdot, \cdot)$. \eqref{eq:gc_bound} is often referred to as generic chaining bound (see Theorem 1.2.6 in \cite{tala05}), and \eqref{eq:gc_conc} is the Theorem 2.2.27 in \cite{tala14}. The functional $\gamma_2(\mathcal{T}, s)$ essentially measures the geometric size of the set $\mathcal{T}$ under the metric $s(\cdot, \cdot)$. To avoid unnecessary complications, we omit the definition of $\gamma_2(\mathcal{T},s)$ here (see Chapter 1 of \cite{tala05} for an introduction if one is interested), 
but provide two of its properties below,
\beq
\gamma_2(\mathcal{T},s_1) \leq \gamma_2(\mathcal{T}, s_2) \ \ \text{if} \ \ s_1(u, v) \leq s_2(u, v), \forall \ u, v \in \mathcal{T}
\eeq
\beq
\gamma_2(\mathcal{T},\eta s) =  \eta \cdot \gamma_2(\mathcal{T},s) \ \ \ \text{for any $\eta > 0$} ~.
\eeq
The important aspect of $\gamma_2$-functional is the following result called \emph{majorizing measure theorem} \cite{tala92,tala05,tala14}. 
\begin{theorem}
\label{theo:mmt}
Given any Gaussian process $\{Y_t\}_{t \in \mathcal{T}}$, define $s(u,v) = \sqrt{\E |Y_u - Y_v |^2}$ for $u, v \in \mathcal{T}$. Then $\gamma_2(\mathcal{T}, s)$ can be upper bounded by
\beq
\gamma_2(\mathcal{T},s) \leq C_0 \E \left[ \sup_{t \in \mathcal{T}} Y_t \right]  ~.
\eeq
\end{theorem}
This theorem is essentially Theorem 2.2.1 in \cite{tala05}. For our purpose, we simply focus on the Gaussian process $\{Y_{\Delta} = \langle \langle G, \Delta \rangle \rangle\}_{\Delta \in \mathcal{A}}$, in which $\mathcal{A} \subseteq \R^{d \times p}$ and $G$ is a standard Gaussian random matrix. Given Theorem \ref{theo:mmt}, the metric $s(U, V) =  \sqrt{\E | \langle \langle G, U -V \rangle \rangle |^2} = \| U -V\|_F$. Therefore we have
\beq
\label{eq:matrix_mmt}
\gamma_2\left(\mathcal{A}, \|\cdot\|_F\right) \leq C_0 \E \left[ \sup_{\Delta \in \mathcal{A}} \langle \langle G, \Delta \rangle \rangle \right] = C_0 w(\mathcal{A})  ~,
\eeq
which will be used in the later proofs. 

\section{Error Bounds with Sub-Gaussian Measurement and Noise}
\label{sec:rand_analysis}
Though the deterministic recovery bound \eqref{eq:gds_bound} in Section \ref{sec:det_analysis} applies to any measurement $\bX$ and noise $\omega$ as long as the assumptions~in \eqref{eq:lambda_cond} and \eqref{eq:re_cond} are satisfied, it is of practical interest to express the bound in terms of the problem parameters, e.g., $d$, $p$ and $n$, for random $\bX$ and $\omega$ sampled from some general and widely used family of distributions. For this work, we assume that $X_i$'s in $\bX$ are i.i.d.~copies of a zero-mean random vector $X$, which is sub-Gaussian with $\vertiii{X}_{\psi_2} \leq \kappa$ for a constant $\kappa$, and the noise $\omega$ contains i.i.d.~centered random variables with $\|\omega_i\|_{\psi_2} \leq \tau$ for a constant $\tau$. In this section, we show that each quantity in \eqref{eq:gds_bound} can be bounded using certain geometric measures associated with the true matrix $\Theta^*$. Further, we show that for unitarily invariant norms, the geometric measures can themselves be bounded in terms of $d$, $p$, $n$, and structures associated with $\Theta^*$.

\subsection{Bounding restricted compatibility constant}
Based on the definition of restricted compatibility constant in \eqref{eq:rcc}, it involves no random quantities and purely depends on $R(\cdot)$ and the geometry of $\mathcal{E}_R(\Theta^*)$. Therefore we directly work on its upper bound for unitarily invariant norms. In general, characterizing the error cone $\mathcal{E}_R(\Theta^*)$ is difficult, especially for non-decomposable $R(\cdot)$. To address this issue, we first need to define the seminorm below.
\begin{defn}
Given two orthogonal subspaces $\mathcal{M}_1, \mathcal{M}_2 \subseteq \R^{d \times p}$ and two vectors $w, z \in \R^d$, the subspace spectral OWL seminorm for $\R^{d \times p}$ is defined as
\beq
\| \Theta \|_{w,z} \triangleq \| \Theta_{\mathcal{M}_1} \|_{w} + \| \Theta_{\mathcal{M}_2} \|_{z} ~,
\eeq
where $\Theta_{\mathcal{M}_1}$ and $\Theta_{\mathcal{M}_2}$ are the orthogonal projections of $\Theta$ onto $\mathcal{M}_1$ and $\mathcal{M}_2$, respectively.
\end{defn}
Next we will construct such a seminorm based on a subgradient $\theta^*$ of the symmetric gauge $f$ associated with $R(\cdot)$ at $\sigma^*$, which can be obtained by solving the so-called \emph{polar operator} \cite{zhys13}
\beq
\theta^* \in \underset{x:f^*(x) \leq 1}{\argmax} \ \left\langle x, \sigma^* \right\rangle ~.
\eeq
Given that $\sigma^*$ is sorted, w.l.o.g.~we may assume that $\theta^*$ is nonnegative and sorted because $\langle \sigma^*, \theta^* \rangle \leq \langle \sigma^*, |\theta^*|^{\downarrow} \rangle$ and $f^*(\theta^*) = f^*(|\theta^*|^{\downarrow})$. Also, we denote by $\theta^*_{\max}$ ($\theta^*_{\min}$) the largest (smallest) element of the $\theta^*$, and define $\rho = \theta^*_{\max} / \theta^*_{\min}$ (if $\theta^*_{\min} = 0$, we define $\rho = + \infty$). Throughout the paper, we will frequently use these notations. As shown in the lemma below, a constructed seminorm based on $\theta^*$ will induce a set $\mathcal{E}'$ that contains $\mathcal{E}_R(\Theta^*)$ and is considerably easier to work with.
\begin{lemm}
\label{lem:superset}
Assume that $\rank(\Theta^*) = r$ and its compact SVD is given by $\Theta^* = U \Sigma V^T$, where $U \in \R^{d \times r}$, $\Sigma \in \R^{r \times r}$ and $V \in \R^{p \times r}$. Let $\theta^*$ be any subgradient of $f(\sigma^*)$, $w = [\theta^*_1, \theta^*_2, \ldots, \theta^*_r, 0, \ldots, 0]^T \in \R^d$, $z = [\theta^*_{r+1}, \theta^*_{r+2}, \ldots, \theta^*_d, 0, \ldots, 0]^T \in \R^d$, $\mathcal{U} = \col(U)$ and $\mathcal{V} = \row(V^T)$, and define $\mathcal{M}_1$, $\mathcal{M}_2$ as
\begin{gather*}
\mathcal{M}_1 = \{ \Theta \ | \ \col(\Theta) \subseteq \mathcal{U}, \row(\Theta) \subseteq \mathcal{V} \} ~, \\
\mathcal{M}_2 = \{ \Theta \ | \ \col(\Theta) \subseteq \mathcal{U}^{\perp}, \row(\Theta) \subseteq \mathcal{V}^{\perp} \} ~,
\end{gather*}
where $\mathcal{U}^{\perp}$, $\mathcal{V}^{\perp}$ are orthogonal complements of $\mathcal{U}$ and $\mathcal{V}$ respectively. Then the specified subspace spectral OWL seminorm $\| \cdot \|_{w,z}$ satisfies
\begin{align*}
\mathcal{E}_R(\Theta^*) \subseteq \mathcal{E}' \triangleq \cone\{ \Delta \ | \ \|\Delta + \Theta^*\|_{w,z} \leq \|\Theta^*\|_{w,z} \}
\end{align*}
\end{lemm}
\proof Both $\mathcal{E}_R(\Theta^*)$ and $\mathcal{E}'$ are induced by scaled (semi)norm balls (i.e., $\Omega_R$ and $\Omega_{w,z}$) centered at $-\Theta^*$, and note that
\begin{align*}
\Theta^*_{\mathcal{M}_1} = \Theta^* ~, \ \ \Theta^*_{\mathcal{M}_2} = 0 ~.
\end{align*}
Thus we obtain
\begin{align*}
\|\Theta^*\|_{w,z} = \| \Theta^*_{\mathcal{M}_1} \|_{w} = \sum_{i=1}^r \sigma^*_i \theta^*_i = \langle \sigma^*, \theta^* \rangle = R(\Theta^*) ~,
\end{align*}
which indicates that the two balls have the same radius. Hence we only need to show that $\|\cdot\|_{w,z} \leq R(\cdot)$. For any $\Delta \in \R^{d \times p}$, assume that the SVD of $\Delta_{\mathcal{M}_1}$ and $\Delta_{\mathcal{M}_2}$ are given by $\Delta_{\mathcal{M}_1} = U_1 \Sigma_1 V_1^T$ and $\Delta_{\mathcal{M}_2} = U_2 \Sigma_2 V_2^T$.
The corresponding vectors of singular values are in the form of $\sigma' = [\sigma'_1, \sigma'_2, \ldots, \sigma'_r, 0, \ldots, 0]^T, \sigma'' = [\sigma''_1, \sigma''_2, \ldots, \sigma''_{d-r}, 0, \ldots, 0]^T \in \R^d$, as $\rank(\Delta_{\mathcal{M}_1}) \leq r$ and $\rank(\Delta_{\mathcal{M}_2}) \leq d-r$. Then we have
\begin{align*}
\|\Delta\|_{w,z} &= \| \Delta_{\mathcal{M}_1} \|_{w} + \| \Delta_{\mathcal{M}_2} \|_{z} = \langle \sigma', w \rangle + \langle \sigma'', z \rangle = \left\langle \theta^*, \left[ \begin{array}{ccc} \sigma'_{1:r} \\ \sigma''_{1:d-r} \end{array} \right] \right\rangle = \langle \langle \Theta, \Delta \rangle \rangle ~,
\end{align*}
where $\Theta = U_1 \Diag(\theta^*_{1:r}) V_1 + U_2 \Diag(\theta^*_{r+1:n}) V_2$. From this construction, we can see that $\theta^*$ are the singular values of $\Theta$, thus $R^*(\Theta) \leq 1$. It follows that
\begin{align*}
\langle \langle \Theta, \Delta \rangle \rangle \leq \max_{R^*(Z) \leq 1} \langle \langle Z , \Delta \rangle \rangle = R(\Delta) ~,
\end{align*}
which completes the proof. \qed

Base on the superset $\mathcal{E}'$, we are able to bound the restricted compatibility constant for unitarily invariant norms by the following theorem.
\begin{theo}
\label{theo:compat}
Assume that there exist $\eta_1$ and $\eta_2$ such that the symmetric gauge $f$ associated with $R(\cdot)$ satisfies
\beq
f(\delta) \leq  \max\left\{  \eta_1 \|\delta\|_1, \ \eta_2 \|\delta\|_2 \right\}
\label{eq:norm_cond}
\eeq
for any $\delta \in \R^{d}$. Then given a rank-$r$ $\Theta^*$, the restricted compatibility constant $\Psi_R(\Theta^*)$ is upper bounded by
\beq
\Psi_R(\Theta^*) \leq 2 \Phi_f(r) + \max \left\{ \eta_2, \eta_1 (1 + \rho) \sqrt{r} \right\} ~,
\eeq
where $\rho = \theta^*_{\max} / \theta^*_{\min}$, and $\Phi_f(r) = \sup_{\|\delta\|_0 \leq r} \frac{f(\delta)}{\|\delta\|_2}$ is called sparse compatibility constant.
\end{theo}
\proof Under the setting of Lemma \ref{lem:superset}, as $\Theta^* \in \mathcal{M}_1$, we have
\begin{gather*}
\|\Delta + \Theta^* \|_{w,z} \leq \|\Theta^*\|_{w,z}  \ \ \ \Longrightarrow \ \ \ \|\Delta_{\mathcal{M}_1} + \Theta^* \|_{w} + \|\Delta_{\mathcal{M}_2}\|_{z}  \leq \|\Theta^*\|_{w} \Longrightarrow  \\
- \|\Delta_{\mathcal{M}_1}\|_{w} + \| \Theta^* \|_{w} + \|\Delta_{\mathcal{M}_2}\|_{z}  \leq \|\Theta^*\|_{w} \ \ \ \Longrightarrow \ \ \ \|\Delta_{\mathcal{M}_2}\|_{z} \leq \|\Delta_{\mathcal{M}_1}\|_{w} ~.
\end{gather*}
As the set $\{ \Delta \ | \ \|\Delta_{\mathcal{M}_2}\|_{z} \leq \|\Delta_{\mathcal{M}_1}\|_{w} \}$ itself is a cone, we obtain
\begin{align*}
\mathcal{E}' \ \subseteq \ \{ \Delta \ | \ \|\Delta_{\mathcal{M}_2}\|_{z} \leq \|\Delta_{\mathcal{M}_1}\|_{w} \}
\end{align*}
Define $\mathcal{M}^{\perp}$ as the orthogonal complement of $\mathcal{M}_1 \oplus \mathcal{M}_2$. By the definition and Lemma \ref{lem:superset}, we have
\begin{align*}
\Psi_R(\Theta^*) &= \sup_{\Delta \in \mathcal{E}_R(\Theta^*)} \frac{R(\Delta)}{\|\Delta\|_{F}} \leq \sup_{\Delta \in \mathcal{E}'} \frac{R(\Delta)}{\|\Delta\|_{F}} \leq \sup_{\|\Delta_{\mathcal{M}_2}\|_{z} \leq \|\Delta_{\mathcal{M}_1}\|_{w}} \frac{R(\Delta)}{\|\Delta\|_{F}} \\
&\leq \sup_{\|\Delta_{\mathcal{M}_2}\|_{z} \leq \|\Delta_{\mathcal{M}_1}\|_{w}} \frac{R(\Delta_{\mathcal{M}^\perp}) + R(\Delta_{\mathcal{M}_1} + \Delta_{\mathcal{M}_2})}{\|\Delta\|_{F}} \\
&\leq \sup_{\Delta \in \mathcal{M}^\perp} \frac{R(\Delta)}{\|\Delta\|_{F}} + \sup_{\frac{\|\Delta_{\mathcal{M}_2}\|_{\tn}}{\|\Delta_{\mathcal{M}_1}\|_{\tn}} \leq \rho} \frac{R(\Delta_{\mathcal{M}_1} + \Delta_{\mathcal{M}_2})}{\|\Delta\|_{F}}
\end{align*}
It is not difficult to see that any $\Delta \in \mathcal{M}^\perp$ has rank at most $2r$, thus
\begin{align*}
\sup_{\Delta \in \mathcal{M}^\perp} \frac{R(\Delta)}{\|\Delta\|_{F}} = \sup_{\Delta \in \mathcal{M}^\perp} \frac{f(\sigma(\Delta))}{\|\sigma(\Delta)\|_{2}} \leq \sup_{\|\delta\|_0 \leq 2r} \frac{f(\delta)}{\|\delta\|_2} \leq 2 \sup_{\|\delta\|_0 \leq r} \frac{f(\delta)}{\|\delta\|_2} = 2 \Phi_f(r) ~.
\end{align*}
Using \eqref{eq:norm_cond} and $\|\Delta_{\mathcal{M}_1} + \Delta_{\mathcal{M}_2}\|_F \leq \|\Delta\|_F$, we have
\begin{align*}
\sup_{\frac{\|\Delta_{\mathcal{M}_2}\|_{\tn}}{\|\Delta_{\mathcal{M}_1}\|_{\tn}} \leq \rho} \frac{R(\Delta_{\mathcal{M}_1} + \Delta_{\mathcal{M}_2})}{\|\Delta\|_{F}} &\leq \sup_{\frac{\|\Delta_{\mathcal{M}_2}\|_{\tn}}{\|\Delta_{\mathcal{M}_1}\|_{\tn}} \leq \rho}  \frac{\max \left\{ \eta_2 \|\Delta\|_F, \ \eta_1 \|\Delta_{\mathcal{M}_1} + \Delta_{\mathcal{M}_2}\|_{\tn} \right\} }{\|\Delta\|_{F}} \\
&\leq \max \left\{ \eta_2, \ \sup_{\Delta \in \mathcal{M}_1}  \frac{ \eta_1(1 + \rho)\|\Delta\|_{\tn}}{\|\Delta\|_F} \right\} \\
&\leq \max \left\{ \eta_2, \eta_1 (1 + \rho) \sqrt{r} \right\} ~,
\end{align*}
where the last inequality uses the fact that any $\Delta \in \mathcal{M}_1$ is at most rank-$r$, and $\|\delta\|_1 \leq \sqrt{r} \|\delta\|_2$ for any $r$-sparse vector $\delta$. Combining all the inequalities, we complete the proof. \qed

\textbf{Remark:} The condition \eqref{eq:norm_cond} might seem cumbersome at the first glance, but the different combinations of $\eta_1$ and $\eta_2$ give us more flexibility. In fact, it trivially encompasses two cases, $\eta_2 = 0$ along with $f(\delta) \leq \eta_1 \|\delta\|_1$ for any $\delta$, and the other way around, $\eta_1 = 0$ along with $f(\delta) \leq \eta_2 \|\delta\|_2$.


\subsection{Bounding restricted convexity $\alpha$}
The condition \eqref{eq:re_cond} is equivalent to
\begin{align*}
\sum_{i=1}^{n} \langle \langle X_i, \Delta \rangle \rangle^2 \geq \alpha > 0, \ \forall \ \Delta \in  \mathcal{E}_R(\Theta^*) \cap \s^{dp-1}  ~.
\end{align*}
In the following theorem, we present the bound for the restricted convexity $\alpha$ in terms of Gaussian width.
\begin{theo}
\label{theo:re}
Assume that $X_i$'s are i.i.d. copies of a centered isotropic sub-Gaussian random matrix $X$ with $\vertiii{X}_{\psi_2} \leq \kappa$, and let $\mathcal{A}_R(\Theta^*) = \mathcal{E}_R(\Theta^*) \cap \s^{dp-1}$. With probability at least $1- \exp(-\zeta w^2(\mathcal{A}_R(\Theta^*)))$, the following inequality holds,
\beq
\label{eq:re}
\inf_{\Delta \in \mathcal{A}}~ \frac{1}{n} \sum_{i=1}^n \langle \langle X_i, \Delta \rangle \rangle^2 ~\geq~ 1 - \xi \frac{\kappa^2 w(\mathcal{A}_R(\Theta^*))}{\sqrt{n}} ~,
\eeq
where $\zeta$ and $\xi$ are absolute constants.
\end{theo}
The proof is essentially an application of generic chaining \cite{tala05,tala14} and the following theorem from \cite{mept07}.
\begin{theo}
There exist absolute constants $c_1$, $c_2$, $c_3$ for which the following
holds. Let $(\Omega,\mu)$ be a probability space, set $H$ be a subset of the unit
sphere of $L_2(\mu)$, i.e., $H \subseteq S_{L_2} = \{ h : \vertiii{h}_{L_2} = 1\}$, and assume that $ \sup_{h \in H}~\vertiii{h}_{\psi_2} \leq \kappa$. Then, for any $\beta> 0$ and $n \geq 1$ satisfying
\beq
\label{eq:beta_cond}
c_1 \kappa \gamma_2(H, \vertiii{\cdot}_{\psi_2}) \leq \beta \sqrt{n}~,
\eeq
with probability at least $1- \exp(-c_2 \beta^2 n/\kappa^4)$,
\beq
\sup_{h \in H}~\left| \frac{1}{n} \sum_{i=1}^n h^2(X_i) - \E\left[h^2\right] \right| \leq \beta~.
\eeq
\label{theo:mpt}
\end{theo}
\noindent{\itshape Proof of Theorem \ref{theo:re}:}\hspace*{1em} For simplicity, we use $\mathcal{A}$ as shorthand for $\mathcal{A}_R(\Theta^*)$. Let $(\Omega, \mu)$ be the probability space that $X$ is defined on, and construct $H = \{ \langle \langle \cdot, \Delta \rangle \rangle \ | \ \Delta \in \mathcal{A} \}$. $\vertiii{X}_{\psi_2} \leq \kappa$ immediately implies that $\sup_{h \in H} \vertiii{h}_{\psi_2} \leq \kappa$. As $X$ is isotropic, i.e., $\E[\langle \langle X, \Delta \rangle \rangle^2] = 1$ for any $\Delta \in \mathcal{A} \subseteq \s^{dp-1}$, thus $H \subseteq S_{L_2}$, and $\E[h^2] = 1$ for any $h \in H$. Given $h_1 = \langle \langle \cdot, \Delta_1 \rangle \rangle, h_2 = \langle \langle \cdot, \Delta_2 \rangle \rangle \in H$, where $\Delta_1, \Delta_2 \in \mathcal{A}$, the metric induced by $\psi_2$ norm satisfies
\begin{align*}
\vertiii{h_1 - h_2}_{\psi_2} = \vertiii{\langle \langle X, \Delta_1 - \Delta_2 \rangle \rangle}_{\psi_2} \leq \kappa \|\Delta_1 - \Delta_2\|_F ~.
\end{align*}
Using the properties of $\gamma_2$-functional and the majorizing measure theorem in Section \ref{sec:prelim}, we have
\begin{align*}
\gamma_2(H, \vertiii{\cdot}_{\psi_2}) ~\leq ~\kappa \gamma_2(\mathcal{A}, \|\cdot\|_F) ~\leq ~ \kappa c_4 w(\mathcal{A})~,
\end{align*}
where $c_4$ is an absolute constant. Hence, by choosing $\beta = c_1 c_4  \frac{\kappa^2 w(\mathcal{A})}{\sqrt{n}}$, we can guarantee that condition \eqref{eq:beta_cond} holds for $H$. Applying Theorem \ref{theo:mpt} to this $H$, with probability at least $1 - \exp(-c_2 c_1^2 c_4^2 w^2(\mathcal{A}))$, we have
\begin{align*}
\sup_{h \in H}~\left| \frac{1}{n} \sum_{i=1}^n h^2(X_i) - 1 \right| \leq \beta ~,
\end{align*}
which implies
\begin{align*}
\inf_{\Delta \in \mathcal{A}} \frac{1}{n} \sum_{i=1}^n \langle \langle X_i, \Delta \rangle \rangle^2 ~  \geq ~ 1 - \beta ~.
\end{align*}
Letting $\zeta = c_2 c_1^2 c_4^2$, $\xi = c_1 c_4$, we complete the proof. \qed

The bound \eqref{eq:re} involves the Gaussian width of set $\mathcal{A}_R(\Theta^*)$, i.e., the error cone intersecting with unit sphere. For unitarily invariant $R(\cdot)$, the following theorem provides a general way to bound $w(\mathcal{A}_R(\Theta^*))$.
\begin{theo}
\label{theo:width}
Under the setting of Lemma \ref{lem:superset}, the Gaussian width $w(\mathcal{A}_R(\Theta^*))$ satisfies
\begin{gather*}
w(\mathcal{A}_R(\Theta^*)) \leq \min \left\{ \sqrt{d p}, \sqrt{ \left( 2  \rho^2 + 1 \right) \left( d + p - r \right) r} \right\} ~.
\end{gather*}
where $\rho = \theta^*_{\max} / \theta^*_{\min}$.
\end{theo}
\proof For simplicity, we again use $\mathcal{A}$ as shorthand for $\mathcal{A}_R(\Theta^*)$. Let $\theta^*$ be any subgradient of $f(\cdot)$ at $\sigma^*$, i.e., $\theta^* \in \partial f(\sigma^*)$, and $\Gamma = U \Diag(\theta^*_{1:r}) V$. Define
\begin{gather*}
\mathcal{D} = \{ W \ | \ W \in \mathcal{M}_2, \ \sigma(W) \preceq z \} ~, \quad \mathcal{K} = \{ \Gamma + W \ | \ W \in \mathcal{D} \} ~,
\end{gather*}
where the symbol ``$\preceq$'' means ``elementwise less than or equal.'' It is not difficult to see that $\mathcal{K}$ is a subset of $\partial R(\Theta^*)$, as any $Z \in \mathcal{K}$ satisfies $R^*(Z) = f^*(\sigma(Z)) \leq f^*(\theta^*) = 1$ and $\langle \langle Z, \Theta^* \rangle \rangle = \langle \sigma(Z), \sigma^* \rangle = \langle \theta^*_{1:r}, \sigma^*_{1:r} \rangle = f(\sigma^*) = R(\Theta^*)$. 
Hence we have
\begin{align*}
\cone(\mathcal{K}) \subset \cone\{\partial R(\Theta^*)\} = \mathcal{N} ~,
\end{align*}
where $\mathcal{N}$ is the polar cone of $\mathcal{E}_R(\Theta^*)$, and the equality follows from the 
Theorem 23.7 of \cite{rock70}. We define the subspace $\mathcal{M}^{\perp}$ as the orthogonal complement of $\mathcal{M}_1 \oplus \mathcal{M}_2$. For the sake of convenience, we denote by $G_1$ ($G_2$, $G_{\perp}$) the orthogonal projection of $G$ onto $\mathcal{M}_1$ ($\mathcal{M}_2$, $\mathcal{M}_{\perp}$), and denote $\cone(\mathcal{K})$ by $\mathcal{C}$. Using \eqref{eq:stat_dim}, we obtain
\beq
\label{eq:cone_1}
\begin{split}
w(\mathcal{A})^2 & \leq \E \left[ \inf_{Z \in \mathcal{N}} \| G - Z\|^2_F \right] \leq \E \left[ \inf_{Z \in \mathcal{C}} \| G - Z\|^2_F \right] \\
&= \E \left[ \inf_{Z \in \mathcal{C}} \| G_1 - Z_1\|^2_F + \| G_2 - Z_2\|^2_F + \| G_{\perp} - Z_{\perp}\|^2_F\right] \\
&= \E \left[ \inf_{\substack{t \geq 0, \\ W \in t \mathcal{D}}} \| G_{1} - t \Gamma\|^2_F + \| G_{2} - W \|^2_F \right] + \E \left[\| G_{\perp} \|^2_F \right]~.
\end{split}
\eeq
To further bound the expectations, we let $t_0 = \frac{\|G_{2}\|_{\op}}{\theta^*_{\min}}$, which is a random quantity depending on $G$. Therefore, we have
\beq
\label{eq:cone_2}
\begin{split}
&\ \ \ \ \ \E  \left[ \inf_{t \geq 0, \ W \in t \mathcal{D}} \| G_{1} - t \Gamma\|^2_F + \| G_{2} - W \|^2_F \right] \\
&\leq \E \left[ \| G_{1} - t_0 \Gamma \|^2_F \right] + \E \left[ \inf_{W \in t_0 \mathcal{D}} \| G_{2} -  W \|^2_F \right] \\
&= \E \left[ \| G_{1} \|^2_F \right] + 2\E \left[ \langle \langle G_{1}, t_0 \Gamma \rangle \rangle \right] + \|\theta^*_{1:r}\|_2^2 \cdot \E \left[ t_0^2 \right] + 0 \\
&= r^2 + 0 + \frac{\|\theta^*_{1:r}\|_2^2}{{\theta^{*2}_{\min}}} \E \left[ \|G_{2}\|_{\op}^2 \right] \\
&\leq r^2 + \frac{\|\theta^*_{1:r}\|_2^2}{{\theta^{*2}_{\min}}} \left[\left(\sqrt{d-r} + \sqrt{p-r}\right)^2 + 2\right] \\
&\leq r^2 + 2 \rho^2 r \left( d + p - 2r \right) ~,
\end{split}
\eeq
where the second equality uses Property 1 and 2 in Section \ref{sec:prelim}, and the second inequality follows from Property 3 and 4. Since $\mathcal{M}_{\perp}$ is a $r(d+p-2r)$-dimensional subspace, by Property 1 we have
\beq
\label{eq:cone_3}
\E \left[ \| G_{\perp} \|^2_F \right] = r(d+p-2r) ~,
\eeq
Combining \eqref{eq:cone_1} \eqref{eq:cone_2} and \eqref{eq:cone_3}, we have
\beq
\label{eq:cone_4}
w(\mathcal{A}) \leq \sqrt{ \left( 2 \rho^2 + 1 \right) \left( d + p - r \right) r} ~.
\eeq
On the other hand, as $\mathcal{A} \subseteq \s^{dp-1}$, we always have
\beq
w(\mathcal{A}) \leq \E \left[\|G\|_{F}\right] \leq \sqrt{\E \left[\|G\|_{F}^2\right]} = \sqrt{d p} ~,
\eeq
which together with \eqref{eq:cone_4} completes the proof. \qed

\subsection{Bounding regularization parameter $\lambda_n$}
In view of Theorem \ref{theo:dantzig}, we should choose the $\lambda_n$ large enough to satisfy \eqref{eq:lambda_cond}, in order for the bound \eqref{eq:gds_bound} to be valid. Hence we need to provide an upper bound for random quantity $R^*\left( \sum_{i=1}^n \omega_i X_i\right)$, which holds with overwhelming probability.
\begin{theo}
Assume that $\bX = \{X_i\}_{i=1}^n$ are i.i.d. copies of a centered isotropic sub-Gaussian random matrix $X$ with $\vertiii{X}_{\psi_2} \leq \kappa$, and the noise $\omega$ consists of i.i.d. centered entries with $\vertiii{\omega_i}_{\psi_2} \leq \tau$. Let $\Omega_R$ be the unit ball of $R(\cdot)$ and $\eta = \sup_{\Delta \in \Omega_R} \|\Delta\|_F$. With probability at least $1 - \exp(- c_1 n) - c_2 \exp\left( - \left(\frac{w(\Omega_R)}{c_3 \eta}\right)^2 \right)$, the following inequality holds
\beq
R^*\left( \sum_{i=1}^n \omega_i X_i\right) \leq c_0 \kappa \tau \sqrt{n} w(\Omega_R)  ~,
\eeq
where $c_0$, $c_1$, $c_2$ and $c_3$ are absolute constants.
\end{theo}
\proof
For each entry in $\omega$, we have
\begin{gather*}
\E[\omega_i^2]^{1/2} \leq \sqrt{2} \vertiii{\omega_i}_{\psi_2} = \sqrt{2} \tau ~,\\
\vertiii{\omega_i^2 - \E[\omega_i^2]}_{\psi_1} \leq 2 \vertiii{\omega_i^2}_{\psi_1} \leq 4 \vertiii{\omega_i}_{\psi_2}^2 \leq 4 \tau^2 ~,
\end{gather*}
where we use the definition of $\psi_2$ norm and its relation to $\psi_1$ norm \cite{vers12}. By Bernstein's inequality, we have
\begin{align*}
\P( \|\omega\|_2^2 - & 2 \tau^2  \geq \epsilon ) \leq \P\left( \|\omega\|_2^2 - \E[\|\omega\|_2^2] \geq \epsilon \right) \leq \exp \left( -c_1 \min \left( \frac{\epsilon^2}{16 \tau^4 n}, \frac{\epsilon}{4 \tau^2} \right) \right) ~.
\end{align*}
Taking $\epsilon = 4\tau^2 n$, we obtain
\beq
\label{eq:lambda_2}
\P\left( \|\omega\|_2 \geq \tau \sqrt{6n} \right) \leq \exp \left( -c_1 n \right) ~.
\eeq
Denote $Y_u = \sum_{i=1}^n u_i X_i$ for $u \in \R^n$. For any $u \in \s^{n-1}$, we have $\vertiii{Y_u}_{\psi_2} \leq c \kappa$ due to
\begin{gather*}
\vertiii{\langle \langle Y_u, \Delta \rangle \rangle}_{\psi_2} = \vertiii{\sum_{i=1}^n u_i \langle \langle X_i, \Delta \rangle \rangle}_{\psi_2} \leq
 c \sqrt{\sum_{i=1}^n u_i^2  \vertiii{\langle \langle X_i, \Delta \rangle \rangle}^2_{\psi_2}} \leq ~ c \kappa \ \ \text{for any $\Delta \in \s^{dp-1}$.}
\end{gather*}
For the rest of the proof, we may drop the subscript of $Y_u$ for convenience. We construct the stochastic process $\{ Z_{\Delta} = \langle \langle Y, \Delta \rangle \rangle \}_{\Delta \in \Omega_R}$, and note that any $Z_U$ and $Z_V$ from this process satisfy,
\begin{gather*}
\P\left( \left|Z_U - Z_V \right| \geq \epsilon \right) = \P\left( \left| \langle \langle Y, U-V \rangle \rangle \right| \geq \epsilon \right)
\leq e \cdot \exp\left( \frac{-C\epsilon^2}{\kappa^2 \|U-V\|_F^2}\right) ~,
\end{gather*}
for some universal constant $C$ due to the sub-Gaussianity of $Y$. As $\Omega_R$ is symmetric, it follows that
\begin{gather*}
\sup_{U, V \in \Omega_R} \left|Z_U - Z_V\right| = 2 \sup_{\Delta \in \Omega_R} Z_{\Delta} ~, \quad \sup_{U, V \in \Omega_R} \|U-V\|_F = 2 \sup_{\Delta \in \Omega_R} \|\Delta\|_F = 2 \eta ~.
\end{gather*}
Let $s(\cdot, \cdot)$ be the metric induced by norm $\kappa \|\cdot\|_F$ and $\mathcal{T} = \Omega_R$. Using \eqref{eq:gc_conc}, we have
\begin{gather*}
\P \left( 2 \sup_{\Delta \in \Omega_R} Z_{\Delta} \geq c_4 \kappa \left( \gamma_2( \Omega_R, \|\cdot\|_F) + \epsilon \cdot 2\eta \right) \right) \leq c_2 \exp \left( - \epsilon^2 \right) ~,
\end{gather*}
where $c_2$ and $c_4$ are absolute constant. By \eqref{eq:matrix_mmt}, there exist constants $c_3$ and $c_5$ such that
\begin{gather*}
\P  (2 R^*(Y) \geq c_5 \kappa \left( w(\Omega_R) + \epsilon \right) ) = \P \left( 2  \sup_{\Delta \in \Omega_R} Z_{\Delta} \geq c_5 \kappa \left( w(\Omega_R) + \epsilon \right) \right) \leq c_2 \exp \left( - \frac{\epsilon^2}{c_3^2 \eta^2}\right) .
\end{gather*}
Letting $\epsilon = w(\Omega_R)$, we have
\beq
\label{eq:lambda_3}
\P  \left( R^*(Y_u) \geq c_5 \kappa w(\Omega_R) \right) \leq c_2 \exp \left( - \left(\frac{w(\Omega_R)}{c_3 \eta}\right)^2  \right)
\eeq
for any $u \in \s^{n-1}$. Combining \eqref{eq:lambda_2},\eqref{eq:lambda_3} and letting $c_0 = \sqrt{6} c_5$, by union bound, we have
\begin{align*}
&\ \ \ \ \ \P  \left(R^*\left( \sum_{i=1}^n \omega_i X_i\right) \geq  c_0 \kappa \tau \sqrt{n} w(\Omega_R) \right) \\
&\leq \P\left(\frac{R^*\left(  Y_{\omega} \right)}{\|\omega\|_2} \geq c_5 \kappa w(\Omega_R) \right) + \P\left( \|\omega\|_2 \geq \tau \sqrt{6n} \right) \\
&\leq  \sup_{u \in \s^{n-1}} \P\left(R^*\left( Y_u \right) \geq c_5 \kappa w(\Omega_R) \right) + \P\left( \|\omega\|_2 \geq \tau \sqrt{6n} \right) \\
&\leq  c_2 \exp\left( - \left(\frac{w(\Omega_R)}{c_3 \eta}\right)^2 \right) + \exp\left(- c_1 n\right) ~,
\end{align*}
which completes the proof. \qed

The theorem above shows that the lower bound of $\lambda_n$ depends on the Gaussian width of the unit ball of $R(\cdot)$. Next we give its general bound for the unitarily invariant matrix norm.
\begin{theo}
\label{theo:ball}
Suppose that the symmetric gauge $f$ associated with $R(\cdot)$ satisfies $f(\cdot) \geq \nu \|\cdot\|_1$. Then the Gaussian width $w(\Omega_R)$ is upper bounded by
\beq
w(\Omega_R) \leq  \frac{\sqrt{d} + \sqrt{p}}{\nu}
\eeq
\end{theo}
\proof As $f(\cdot) \geq \nu \|\cdot\|_1$, we have
\begin{align*}
R(\cdot) \geq \nu \|\cdot\|_{\tn} \ \ \ \ \Longrightarrow \ \ \ \ \Omega_R \subseteq \Omega_{\nu \|\cdot\|_{\tn}} ~.
\end{align*}
Hence it follows that
\begin{align*}
w\left(\Omega_R\right) \leq w\left(\Omega_{\nu \|\cdot\|_{\tn}}\right) = \frac{w\left(\Omega_{\|\cdot\|_{\tn}}\right)}{\nu} = \frac{\E \|G\|_{\op}}{\nu} \leq \frac{\sqrt{d} + \sqrt{p}}{\nu} ~,
\end{align*}
where the last inequality follows from the Property 4 of Gaussian random matrix.  \qed

\section{Examples}
\label{sec:example}
In Section \ref{sec:rand_analysis}, we observe that the matrix recovery error is determined by the three geometric measures of sets associated with the true matrix $\Theta^*$, respectively given by $\Psi_R(\Theta^*)$, $w(\mathcal{A}_R(\Theta^*))$ and $w(\Omega_R)$. Combining those results, we note that if the number of measurements $n > O(w^2(\mathcal{A}_R(\Theta^*)))$, then the recovery error, with high probability, is upper bounded by
\beq
\|\hat{\Theta} - \Theta^*\|_F \leq O\left( \Psi_R(\Theta^*) \cdot \frac{w(\Omega_R)}{\sqrt{n}} \right) ~.
\eeq
In this section, we give two examples based on the trace norm \cite{capl09} 
and the recently proposed spectral $k$-support norm \cite{mcps14} 
to illustrate how to bound these geometric measures and obtain bounds on the estimation error.

\subsection{Trace norm}
Trace norm has been widely used in low-rank matrix recovery. The trace norm of $\Theta^*$ is basically the $\ell_1$ norm of $\sigma^*$, i.e., $f = \|\cdot\|_1$. Now we turn to the three geometric measures. Assuming that $\rank(\Theta^*) = r \ll d$, one subgradient of $\|\sigma^*\|_1$ is $\theta^* = [1, 1, \ldots, 1]^T$.

\textbf{Restricted compatibility constant $\Psi_{\tn}(\Theta^*)$:} It is obvious that assumption \eqref{eq:norm_cond} will hold for $f$ by choosing $\eta_1 = 1$ and $\eta_2 = 0$, and we have $\rho = 1$. The sparse compatibility constant $\Phi_{\ell_1}(r)$  is $\sqrt{r}$ because $\|\delta\|_1 \leq \sqrt{r} \|\delta\|_2$ for any $r$-sparse $\delta$. Using Theorem \ref{theo:compat}, we have
\begin{align*}
\Psi_{\tn}(\Theta^*) \leq 4 \sqrt{r}~.
\end{align*}
\textbf{Gaussian width $w(\mathcal{A}_{\tn}(\Theta^*))$:} As $\rho = 1$, it follows immediately from Theorem \ref{theo:width} that 
\begin{align*}
w(\mathcal{A}_{\tn}(\Theta^*)) \leq \sqrt{3r(d+p-r)}~.
\end{align*}
\textbf{Gaussian width $w(\Omega_{\tn})$:} Using Theorem \ref{theo:ball} and noting $\nu = 1$, it is easy to see that
\begin{align*}
w(\Omega_{\tn}) \leq \sqrt{d} + \sqrt{p}~.
\end{align*}
Putting all the results together, we can conclude that when $n > O(r(d+p-r))$, the recovery error of rank-$r$ $\Theta^*$ using trace norm, with high probability, satisfies
\begin{align}
\|\hat{\Theta} - \Theta^*\|_F \leq O\left(\sqrt{\frac{rd}{n}} + \sqrt{\frac{rp}{n}}\right) ~,
\end{align}
which matches the bound in \cite{care09}.
\subsection{Spectral $k$-support norm}
The $k$-support norm proposed in \cite{arfs12} is defined as
\begin{align*}
\|\theta\|_k^{sp} \triangleq \inf_{\sum_{i} u_i = \theta } \Big\{ \sum_{i}\|u_i\|_2 ~\Big |~ \|u_i\|_0 \leq k  \Big\} ~,
\end{align*}
and its dual norm is simply given by 
\begin{align*}
\|\theta\|_k^{sp*} = \| |\theta|^{\downarrow}_{1:k} \|_2 ~.
\end{align*}
It is shown that $k$-support norm has similar behavior as elastic-net regularizer \cite{zoha05}. Spectral $k$-support norm (denoted by $\|\cdot\|_{\sk}$) of $\Theta^*$ is defined by applying the $k$-support norm on $\sigma^*$, i.e., $f = \|\cdot\|_{k}^{sp}$, which has demonstrated better performance than trace norm in matrix completion task \cite{mcps14}. For simplicity, We assume that $\rank(\Theta^*) = r = k$ and $\|\sigma^*\|_2 = 1$. One subgradient of $\| \sigma^* \|_{k}^{sp}$ can be $\theta^* = \left[ \sigma^*_1, \sigma^*_2, \ldots, \sigma^*_r, \sigma^*_r, \ldots, \sigma^*_r \right]^T$.

\textbf{Restricted compatibility constant $\Psi_{\sk}(\Theta^*)$:} The following relation has been shown for $k$-support norm in \cite{arfs12},
\beq
\label{eq:ksup_en}
\max \left\{ \|\cdot\|_2, \frac{\|\cdot\|_1}{\sqrt{k}} \right\} \leq \|\cdot\|_{k}^{sp} \leq \sqrt{2} \max \left\{ \|\cdot\|_2, \frac{\|\cdot\|_1}{\sqrt{k}} \right\} .
\eeq
Hence the assumption \eqref{eq:norm_cond} will hold for $\eta_1 = \sqrt{\frac{2}{k}}$ and $\eta_2 = \sqrt{2}$, and we have $\rho = \frac{\sigma^*_1}{\sigma^*_r}$. The sparse compatibility constant $\Phi_k^{sp}(r) = \Phi_k^{sp}(k) = 1$ because $\|\delta\|_k^{sp} = \|\delta\|_2$ for any $k$-sparse $\delta$. Using Theorem \ref{theo:compat}, we have
\begin{align*}
\Psi_{\sk}(\Theta^*) \leq 2 \sqrt{2} + \sqrt{2} \left(1+\frac{\sigma^*_1}{\sigma^*_r}\right) = \sqrt{2} \left( 3 + \frac{\sigma^*_1}{\sigma^*_r} \right)~.
\end{align*}
\textbf{Gaussian width $w(\mathcal{A}_{\sk}(\Theta^*))$:} we note that $\rho = \frac{\sigma^*_1}{\sigma^*_r}$, and Theorem \ref{theo:width} implies that 
\begin{align*}
w(\mathcal{A}_{\sk}(\Theta^*)) \leq \sqrt{r(d+p-r)\left[\frac{2 \sigma^{*2}_1}{\sigma^{*2}_r} + 1\right]}~.
\end{align*}
\textbf{Gaussian width $w(\Omega_{\sk})$:} lower bound of \eqref{eq:ksup_en} implies that $\nu = 1 / \sqrt{k} = 1 / \sqrt{r}$. By Theorem \ref{theo:ball}, we get
\begin{align*}
w(\Omega_{\sk}) \leq \sqrt{r} (\sqrt{d} + \sqrt{p})~.
\end{align*}
Given the upper bounds for geometric measures, we can conclude that when $n > O(r(d+p-r))$, the recovery error of rank-$r$ $\Theta^*$ using spectral $k$-support norm, with high probability, satisfies
\begin{align}
\label{eq:ksup_bound}
\|\hat{\Theta} - \Theta^*\|_F \leq O\left(\sqrt{\frac{rd}{n}} + \sqrt{\frac{rp}{n}}\right) ~.
\end{align}
The spectral $k$-support norm was first introduced in \cite{mcps14}, in which no statistical results are provided. Although \cite{gubg15} investigated the statistical aspects of spectral $k$-support norm in matrix completion setting, the analysis was quite different from our setting. Hence the error bound \eqref{eq:ksup_bound} is new in the literature.

\section{Conclusions}
\label{sec:conc}

In this work, we present the recovery analysis for matrices with general structures, under the setting of sub-Gaussian measurement and noise. Base on generic chaining and Gaussian width, the recovery guarantees can be succinctly summarized in terms of some geometric measures. For the class of unitarily invariant norms, we also provide general bounds of these measures, which can significantly facilitate the analysis.


\vspace*{3mm}
{\bf Acknowledgements:} The research was supported by NSF grants IIS-1447566, IIS-1422557, CCF-1451986, CNS-1314560, IIS-0953274, IIS-1029711, and by NASA grant NNX12AQ39A.


\bibliography{ref}
\bibliographystyle{plain}

\end{document}